\documentclass[a4paper,11pt]{article}
\usepackage{hyperref}
\hypersetup{colorlinks = true,
            linkcolor = magenta,
            urlcolor  = blue,
            citecolor = green,
            anchorcolor = blue}

\usepackage{amsmath}
\usepackage{amssymb}
\usepackage{times}
\usepackage{caption}
\usepackage{epsfig}
\usepackage{float}
\usepackage{graphicx}
\usepackage{authblk} 
\usepackage{subfigure}
\usepackage{times}
\usepackage{comment}
\usepackage{placeins}

\usepackage[utf8]{inputenc}
\usepackage[numbers,sort&compress]{natbib}
\usepackage[top=2.5cm, bottom=2.5cm, left=2.5cm, right=2.5cm]{geometry}
\begin{document}
\title{Adversarial Robustness by Design  through\\ Analog Computing and Synthetic Gradients}

\author[1]{Alessandro Cappelli\thanks{Equal contribution. Corresponding authors: \texttt{\{alessandro,ruben\}@lighton.ai}}}
\author[1,2]{Ruben Ohana$^*$}
\author[1,2]{Julien Launay}
\author[3,4]{Laurent Meunier}
\author[1]{Iacopo Poli}
\author[1,2,5]{Florent Krzakala}
\affil[1]{LightOn, Paris, France}
\affil[2]{Laboratoire de Physique de l’École Normale Supérieure, Université
  PSL, CNRS, \protect \\ Sorbonne Université, Université Paris-Diderot, Sorbonne Paris
  Cité, Paris, France}
\affil[3]{Facebook AI Research, Paris, France}
\affil[4]{Université Paris-Dauphine, PSL Research University, CNRS, LAMSADE, Paris, France}
\affil[5]{IdePHICS Laboratory, EPFL, Switzerland}
\date{}

\maketitle

\begin{abstract}
\vspace{-0.3cm}
We propose a new defense mechanism against adversarial attacks inspired by an optical co-processor, providing robustness without compromising natural accuracy in both white-box and black-box settings. This hardware co-processor performs a nonlinear fixed random transformation, where the parameters are unknown and impossible to retrieve with sufficient precision for large enough dimensions. In the white-box setting, our defense works by obfuscating the parameters of the random projection. Unlike other defenses relying on obfuscated gradients, we find we are unable to build a reliable backward differentiable approximation for obfuscated parameters. Moreover, while our model reaches a good natural accuracy with a hybrid backpropagation - synthetic gradient method, the same approach is suboptimal if employed to generate adversarial examples. We find the combination of a random projection and binarization in the optical system also improves robustness against various types of black-box attacks. Finally, our hybrid training method builds robust features against transfer attacks. We demonstrate our approach on a VGG-like architecture, placing the defense on top of the convolutional features, on CIFAR-10 and CIFAR-100. Code is available at \url{https://github.com/lightonai/adversarial-robustness-by-design}.

\end{abstract}

\section{Introduction}

Neural networks are sensitive to small, imperceptible to humans, perturbations of their inputs that can cause state-of-the-art classifiers to completely fail \cite{goodfellow2014explaining}. As deep learning models are deployed in real-world applications, guaranteeing their robustness to malicious actors becomes increasingly important: for instance, an adversarial image could evade automated content filtering on social networks~\cite{garcelon2020adversarial}.
Adversarial attacks can be carried out in different frameworks: in the white-box setting, the attacker has full access to the model, while black-box attacks only rely on queries. It is also possible to craft an attack on a different model and transfer it to the model targeted~\cite{papernot2016transferability}. There is no universal defense, and state-of-the-art techniques often come with a large computational cost, as well as reduced natural accuracy~\cite{tsipras2018robustness}.
Some of these defenses rely on obfuscated gradients: the model is designed so that the gradients are unsuitable for attacks, for instance by using non-differentiable layers. However, attackers can choose to alter the network structure, using Backward Pass Differentiable Approximation (BPDA)~\cite{obfuscated-gradients}, replacing obfuscating layers with well-behaved approximations. Furthermore, approaches relying on obfuscation do not generally provide robustness against transfer and black-box attacks.


We expand the idea of obfuscated gradients to \emph{obfuscated parameters}: we physically implement a fixed random projection followed by a non-linearity using an optical co-processor, where only the distribution of the random matrix entries is known and not their values. Even though retrieval is possible, the computational cost becomes quickly prohibitive with increasing dimension, and is limited in precision~\cite{gupta2019don,gupta2020fast}.
To train layers below our non-differentiable defense, we draw inspiration from Direct Feedback Alignment (DFA) \cite{nokland2016direct} and  bypass it in the backward pass. We use a random mapping of the global error to train the layer below the co-processor, while the layers further downstream perform backpropagation (BP) from this \textit{synthetic gradient} signal. This comes at no natural accuracy cost.

Our defense is robust by design against white-box attacks: we find that BPDA is ineffective against obfuscated parameters, and attackers are forced to rely on DFA to attack the network. We develop such DFA attacks, and find them much less effective than attacks based on backpropagation on BP-trained networks, confirming results from~\cite{akrout2019adversarial}. 

We also test models incorporating our defense against black-box and transfer attacks, and find that they are more robust than their vanilla counterparts. As parameter obfuscation alone cannot explain robustness in these settings, we perform an ablation study to identify the mechanism responsible for it. For black-box attacks, the combination of binarization and random projection given by the optical co-processor gives a clear contribution to robustness. For transfer attacks, we find the training method itself produces more robust features. Overall, our hybrid training method in conjunction with our optical layer provide a complete defense.
\begin{figure}[t]
    \centering
    \includegraphics[width=\linewidth]{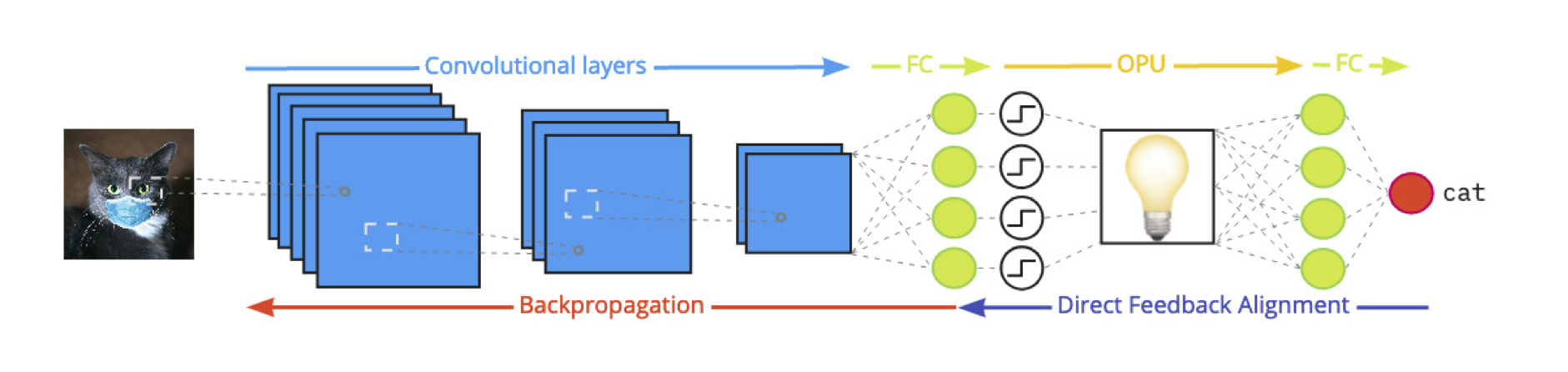}
  \caption{A convolutional neural network with an Optical Processing Unit (OPU), our analog defense layer against adversarial attacks. All together, the unknown parameters of the analog operation, the binarization, and the hybrid training method based on Direct Feedback Alignment (DFA) form a defense against white-box, black-box, and transfer attacks.
}
  \label{fig: vgg-opu}
\end{figure}
\paragraph{Contributions.} Our study extends gradient obfuscation defenses, introducing a defense based on obfuscation of parameters: this obfuscation is guaranteed by the physical implementation of the defensive layer using an optical co-processor. In the process of evaluating it against black-box and transfer attacks, we find that additional robustness components arise from the binarization and the hybrid training method we use to bypass our non-differentiable defense. In summary:
\begin{itemize}
    \item We introduce a defense by obfuscated parameters (Section \ref{section:preliminaries}), implemented by an optical co-processor and develop a hybrid training method inspired by DFA to train models incorporating our defense. This defense comes at no cost in natural accuracy, and no computational cost as all relevant computations are off-loaded to the optical co-processor.
    \item Our defense is robust by design against FGSM and PGD white-box attacks (Section \ref{subsec:white}). To perform white-box attacks against our defense, we introduce hybrid DFA attacks able to bypass our obfuscation, in the spirit of BPDA. However, these are way less effective in creating compelling adversarial examples.
    \item We find our defense robust against a variety of black-box attacks (NES, bandits and parsimonious). In particular, our defense is significantly more resistant to parsimonious attacks, the strongest of the three evaluated (Section \ref{subsec:black}). Our defense also provides robustness against transfer attacks (Section \ref{subsec:transfer}).
    \item Since parameters obfuscation cannot explain black-box and transfer robustness, we perform an ablation study to understand the mechanism behind it (Section \ref{section:ablation}). We show that all the elements of our defense are necessary to get robustness to black-box attacks. Surprisingly, for transfer attacks, we find that our training method alone brings robustness.
\end{itemize}
We perform all experimental benchmarks on CIFAR-10, and CIFAR-100 \cite{krizhevsky2009cifar}. White-box and transfer of attacks results are obtained with a real optical co-processor as a defense whereas for black-box attacks, a simulated co-processor is used for convenience.
\section{Preliminaries}\label{section:preliminaries}
\subsection{Adversarial attacks and defenses}
\paragraph{White-box attacks.} White-box attacks are adversarial attacks where the attacker is assumed to have full access to the model, including its parameters. In this case, the attacker usually computes a \textit{gradient attack} (e.g. FGSM~\cite{goodfellow2014explaining}, PGD~\cite{kurakin2016adversarial,madry2017towards}, or Carlini \& Wagner~\cite{carlini2017towards}). These attacks are often fast, effective, and easy to compute. Some defenses obfuscate the gradients to neutralize these attacks: however, it is often possible to build a differentiable approximation (BPDA) to perform gradient-based attacks~\cite{obfuscated-gradients}, and black-box attacks entirely elude such defenses. In our work, we focus on PGD for white-box scenarios (see Figure \ref{figsup: WB attack result} in the supplementary for FGSM results). 
\vspace{-0.35cm}
\paragraph{Black-box attacks.} The black-box setting assumes that the attacker has only limited access to the network: for instance, only the label, or the logits for a given input are known. There exist two main categories of black-box attacks. On one hand, gradient estimation attacks~\cite{chen2017zoo,ilyas2018black,ilyas2018prior} aim to estimate the gradient of the loss with respect to the input to mimic gradient-based attacks. On the other hand, adversarial attacks are transferable~\cite{papernot2016transferability}: an attack effective on a given network is likely to also fool another network. More recently, black-box methods based on optimisation tools derived from genetic algorithms~\cite{meunier2019yet,andriushchenko2019square} and combinatorial optimization~\cite{moon2019parsimonious} have been introduced. We evaluate our defense against NES and bandits, two gradient-estimation attacks, and parsimonious black-box methods, as well as transfer of attacks.

\paragraph{Defenses} Historically, the first defense proposed against attacks was adversarial training~\cite{goodfellow2014explaining,madry2017towards} (i.e. training the neural network through a min-max optimization framework) thus including adversarial robustness as an explicit training objective. Despite its simplicity and lack of theoretical guarantees, adversarial training is still one of the most effective defense against adversarial examples. Theoretically proven defenses also exist, such as randomized smoothing~\cite{lecuyer2018certified,KolterRandomizedSmoothing,pinot2019theoretical,araujo2020} or convex relaxation~\cite{wong2018provable,wong2018scaling}. In the literature, numerous defenses do not evaluate their models on attacks adapted to the defense~\cite{tramer2020adaptive}: especially in the case of gradient obfuscation~\cite{obfuscated-gradients}, which can result in a false sense of security. In contrast, we evaluate our defense in a wide range of scenarios, and introduce new DFA-based white-box attacks. We also drive a thorough and rigorous ablation study, to better understand the mechanisms underlying the robustness of our defense.\\
Finally, defense techniques often demand extra computations and reduce natural accuracy. Instead, our defense computations are offloaded to the optical co-processor and the decrease in natural accuracy is minimal, in contrast with adversarial training.

\subsection{Our defense: the Optical Processing Unit}\label{subsec:opu}
Our defense relies on an analog layer implemented by an Optical Processing Unit (OPU)\footnote{Accessible through LightOn Cloud \url{https://cloud.lighton.ai/}}.
The OPU is a co-processor that multiplies an input vector $x$ by a fixed random matrix $U$, using light scattering through a diffusive medium~\cite{lighton2020}. The measurement process implies an inherent absolute value squared non-linearity. Effectively, the operation performed is:
\begin{equation}
    m = \lvert Ux\rvert^2
    \label{eq: opu operation}
\end{equation}

The OPU input is binary (1 bit) and its output is encoded in 8 bits. We thereafter refer to OPU as the combination of the binarization operation and the matrix product. The size of the random matrix can reach $10^6\times 10^6$, and its entries are complex Gaussian distributed, but their values are not known. While our defense can be simulated without dedicated hardware, an advantage of using the OPU is that even if the host system is compromised, the random matrix remains unknown, as it is physically implemented by the diffusive medium of the OPU. 
\begin{figure}[!b]
    \centering
    \includegraphics[width=0.7\linewidth]{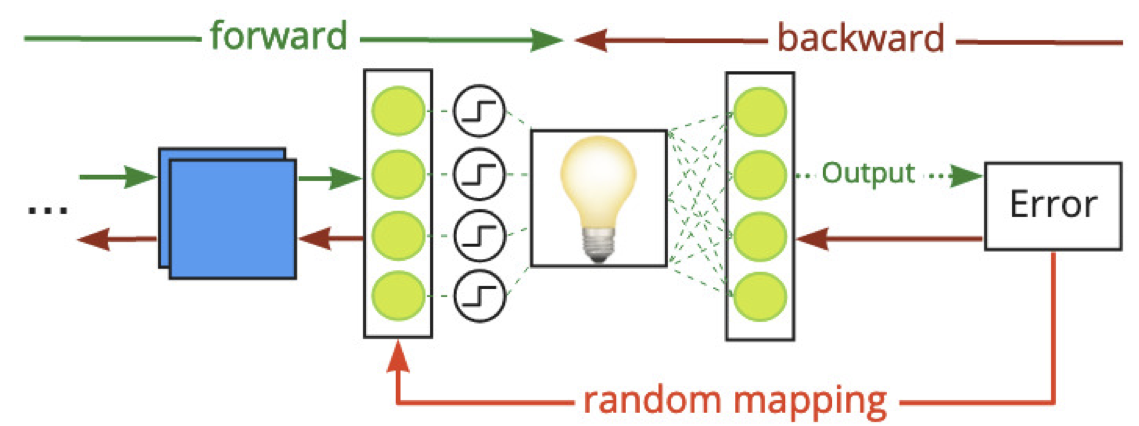}
    \caption{The non-differentiable defense layer is bypassed in the backward path by using a random mapping of the error as a synthetic gradient. This signal is then used in the backpropagation update of the previous layers.}
    \label{fig:hybrid-training}
\end{figure}

Retrieval of the matrix is possible \cite{gupta2019don,gupta2020fast} with direct access to the OPU, but it is computationally expensive for large enough dimensions and relative errors can be as high as $30\%$. The fastest known method~\cite{gupta2020fast} for the  retrieval of a matrix $U \in \mathbb{C}^{M\times N}$  relies on the multilateration of anchor signals and has $O(MN\log N)$ time complexity. For $N\sim10^4$ and $M\sim10^5$, retrieval with relative error of $32.0\%$ takes 72 minutes. If we wanted to recover the same matrix with a relative error of $8.0\%$, we would need 19 hours, as decreasing the relative error has a quadratic cost -- see section \ref{sectionsup: retrieval} in the supplementary for details. The optical system can scale up to $N, M \sim 10^6$: at these dimensions the random matrix alone takes about $8\mbox{ TB}$ to store, making memory the main constraint in the retrieval. Finally, it is also easily possible to change the entries of the matrix of the optical system to another draw from the same probability distribution. To adapt to this new random matrix, only the classifier has to be fine-tuned: this enables a defense strategy where the random matrix is regularly resampled, preventing malicious actors from having enough time to recover the it.

Accordingly, our defense effectively achieves \emph{parameter obfuscation}, preventing attackers from building a differentiable approximation that can be used to reliably generate adversarial examples.
In our work we use an actual optical co-processor for white-box and transfer attacks, and a simulated one for black-box. Note that while we simulate input binarization, we don't simulate output quantization to 8 bits, as we find this quantization to be of little influence.   

\subsection{Network training and adversarial examples generation with synthetic gradients}\label{subsec:training}
Because its parameters are obfuscated, it is not possible to perform backpropagation through the layer implemented by the optical co-processor. To train neural networks incorporating our defense, we draw inspiration from Direct Feedback Alignment \cite{nokland2016direct}, and design a hybrid training method (Figure \ref{fig:hybrid-training}). We train the layers above our defense through backpropagation, but train the layer below it by directly using a random projection of the global error as the teaching signal. To account for the inability of DFA to train convolutional layers \cite{launay2019principled}, the layer directly before our defense should be a fully-connected layer. This synthetic signal is then backpropagated to convolutional layers further down. We use the same hybrid method to generate white-box attacks against our networks.

\section{Experimental results}
We place our defense after the convolutional layers of a VGG-16 architecture \cite{simonyan2014very}. We call this network \textit{VGG-OPU}. The specific hyperparameters of our model are precised in section \ref{sectionsup: hyperparameters} of the supplementary. The training is performed with the hybrid BP-DFA algorithm discussed in the Section \ref{subsec:training}, and shown in Figure \ref{fig: vgg-opu}. The code\footnote{\url{https://github.com/lightonai/dfa-scales-to-modern-deep-learning}} of DFA is taken from~\cite{launay2020direct}. We consider the CIFAR-10 and CIFAR-100 datasets.

We attack our models using only images that were correctly classified, with the exception of white-box attacks, where we use the full datasets. All the attacks are \textit{untargeted}: we aim to change the classification without any specific target label. Finally, the loss used for computing the attacks is the cross-entropy between the output of the classifier for a given input and its label: $l(x,y) = -\log p_\theta(y|x)$. As we consider untargeted attacks, we aim to maximize it.

\subsection{White-box attacks}\label{subsec:white}

We first consider white-box attacks: the attacker has full knowledge of the model and its parameters, and can craft adversarial examples by gradient-based methods. We show that our parameter obfuscation approach makes these attacks significantly less effective, forcing them to rely on imprecise gradient approximations based on DFA or BPDA.

\paragraph{Background} We perform PGD attacks~\cite{madry2018towards}, where the adversarial example $x$ is iteratively optimized with: 
\begin{align*}
    x^{t+1} = \Pi_{B_\infty(x,\epsilon)}\left[x^t+\alpha\text{sign}\left(\nabla_x l(x^t,y)\right)\right]
\end{align*}
where $\Pi_{B_\infty(x,\epsilon)}$ is the orthogonal projection on $B_\infty(x,\epsilon):=\{x':\lVert x'-x\rVert_\infty\leq \epsilon\}$ and $x^0=x$. The quantity $\epsilon$ is the strength of the perturbation. In the particular case of a single iteration, the attack is called FGSM~\cite{goodfellow2014explaining} -- as FGSM is weaker than PGD, we focus on PGD in this section, and leave FGSM results to the supplementary (Figure \ref{figsup: WB attack result}). 

To attack networks despite our non-differentiable defense, we either use our hybrid training method to generate attack gradients by skipping our defense, or consider building a BPDA of our defense: the sign function is approximated by a $\tanh$, and the unknown random weights are approximated by another set of random weights. 
 \begin{figure}[!h]
    \centering
    \includegraphics[width=0.95\linewidth]{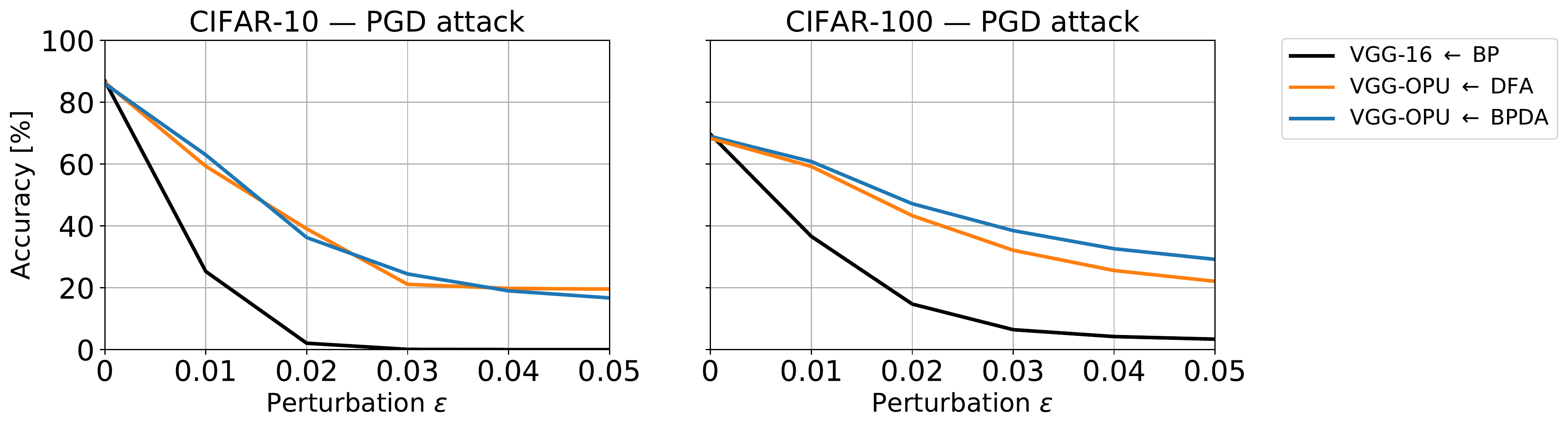}
    \caption{Notation: $<$model$>$ $\leftarrow$ $<$attack gradients$>$. For example VGG $\leftarrow$ BP means that a VGG-16 is attacked with gradients computed with backpropagation. The VGG-OPU model is systematically more robust than the VGG-16 baseline. The failure of BPDA to produce better attacks than our hybrid training method (here abbreviated DFA) confirms that parameters obfuscation enables robustness to white-box attacks by design.}
    \label{fig: WB attack result}
\end{figure}
\vspace{-0.2cm}
\paragraph{Results} Results are shown in Figure \ref{fig: WB attack result}. We find the VGG-OPU model incorporating our defense performs better than the VGG-16 baseline for any value of $\epsilon$, with gains in accuracy under attack ranging from $20\%$ to $40\%$. For the largest values of $\epsilon$ considered, while the accuracy of the baseline goes to zero, the VGG-OPU model is still performing better than a random guess. These results also show that our obfuscated parameters approach cannot be fooled by a simple BPDA like obfuscated gradients can be: BPDA is here ineffective at finding better attacks than simply bypassing our defense with DFA.
Finally, we note the increased robustness does not come at a natural accuracy cost ($\epsilon=0$).

\subsection{Black-box attacks}
If the obfuscated parameters approach provides robustness by design against white-box attacks, black-box approaches should be not affected, since they do not require knowledge of the weights. However, we find our defense still brings robustness against such attacks.
\subsubsection{NES, bandits, and parsimonious attacks}\label{subsec:black}
\paragraph{Background} To further test the robustness of our defense, we perform strong black-box attacks: NES ~\cite{ilyas2018prior}, bandits~\cite{ilyas2018prior}, and parsimonious attacks~\cite{moon19aparsimonous}.

NES and bandits attacks consist in efficiently estimating the gradient of the loss with regards to the input $x$: 
\begin{align*}
    \nabla_x l(x,y)&\simeq \nabla_x \mathbb{E}_{n\sim\mathcal{N}(x,\sigma^2I)}\left[l(n,y)\right]\\
    &\simeq  \frac{1}{\sigma N}\sum_{i=1}^N  \delta_i l(x+\sigma \delta_i,y)
\end{align*}
where $y$ is the true label, $\delta_i$ are standard Gaussian random variables and $N$ the sample size of the Monte-Carlo estimation.  Bandits attacks are an extension of NES attacks, taking advantage of gradients correlations for close pixels and between gradient steps.
\begin{figure}[!b]
    \includegraphics[width=0.95\linewidth]{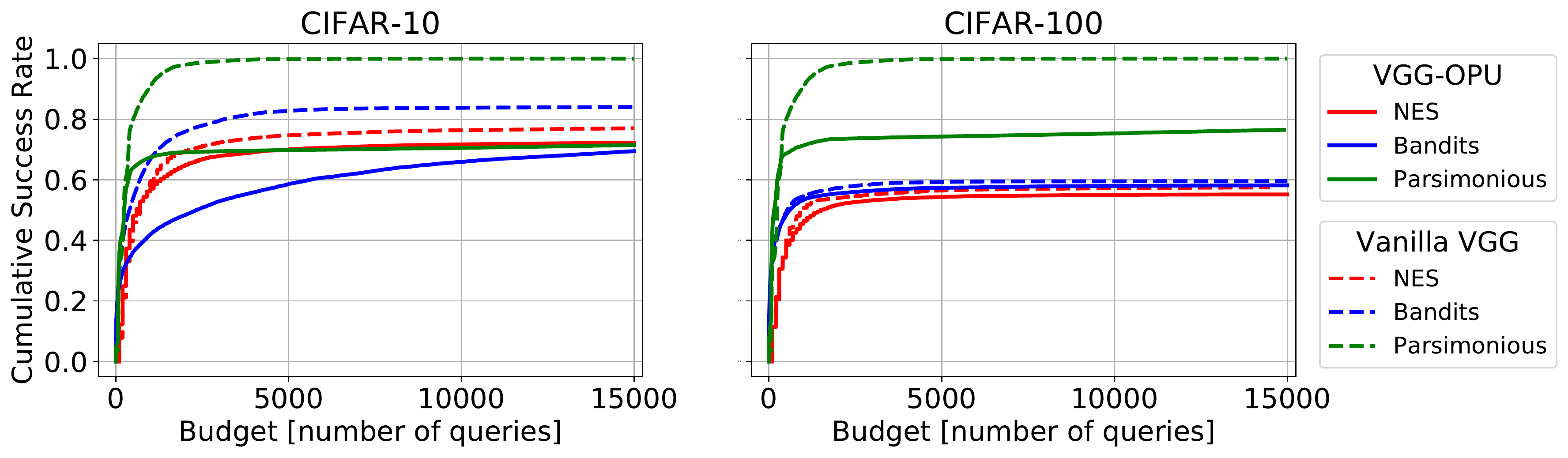}
    \caption{Cumulative success rate with respect to the number of queries for different black-box attacks on the CIFAR-10 and CIFAR-100 datasets, on a VGG-16 \textbf{(dashed lines)} and a VGG-OPU with a simulated OPU \textbf{(plain lines)}. The architectures with our defense perform on par or better than the baseline. The parsimonious attack has the highest success rate, that with our defense drops by 0.3 on CIFAR-10 and 0.2 on CIFAR-100.}
    \label{fig: BB NES bandit}
\end{figure}
To ensure we are not robust only against gradient estimation-based black-box attacks, we also consider parsimonious attacks. This attack is different in nature from the previous two, as it is inspired by combinatorial optimization. It is worth mentioning that this attack is not very sensible to hyperparameters choice~\cite{moon19aparsimonous}. Parsimonious attacks are also significantly stronger, outperforming NES and bandits approaches.  

We measure the Cumulative Success Rates (CSR) in terms of elapsed queries budget. We fix a maximum budget of $15000$ queries to the classifier for black-box attacks. This budget is sufficient to reach a plateau in the success rate of the attacks, and to achieve $100\%$ success rate with parsimonious attacks on an undefended baseline. We select attack hyperparameters obtaining the best success rates, as described in section \ref{sectionsup: hyperparameters} of the supplementary.
\paragraph{Results} Results are shown in Figure \ref{fig: BB NES bandit}. For the gradient-estimation attacks (NES and bandits), our defense improves robustness with respect to the baseline by decreasing the CSR respectively by $5\%$ and 10 $\%$ for the largest budget on CIFAR-10. On CIFAR-100, the improvement in robustness is minimal, of the order of $1\%$ for both attacks. However, our defense shows significant improvement against parsimonious attacks (that largely outperform both NES and bandits), reducing the CSR by $30\%$  on CIFAR-10 and by $24\%$ CIFAR-100. Overall, the best attack on Vanilla VGG reaches $100\%$ CSR on both datasets whereas the best attack on a VGG with our defense reaches only $70\%$ on CIFAR-10 and $76\%$ on CIFAR-100.

\subsubsection{Transfer attacks}\label{subsec:transfer}
\paragraph{Background} Finally, we test the robustness of our defense against transfer attacks. In this scenario, attacks are crafted on a separate source network built by the attacker, similar to the target network. This is a form of black-box attack that does not require queries to the target network.

To evaluate transfer attacks, we first create a test set of well classified samples common to both the source and target network. We then perform attacks on a VGG-16 model on this dataset using the PGD algorithm, building a collection of adversarial examples to transfer to other networks -- FGSM results are presented in Figure \ref{figsup:transfer result} of the supplementary.
\begin{figure}[b!]
    \centering
    \includegraphics[width=0.95\linewidth]{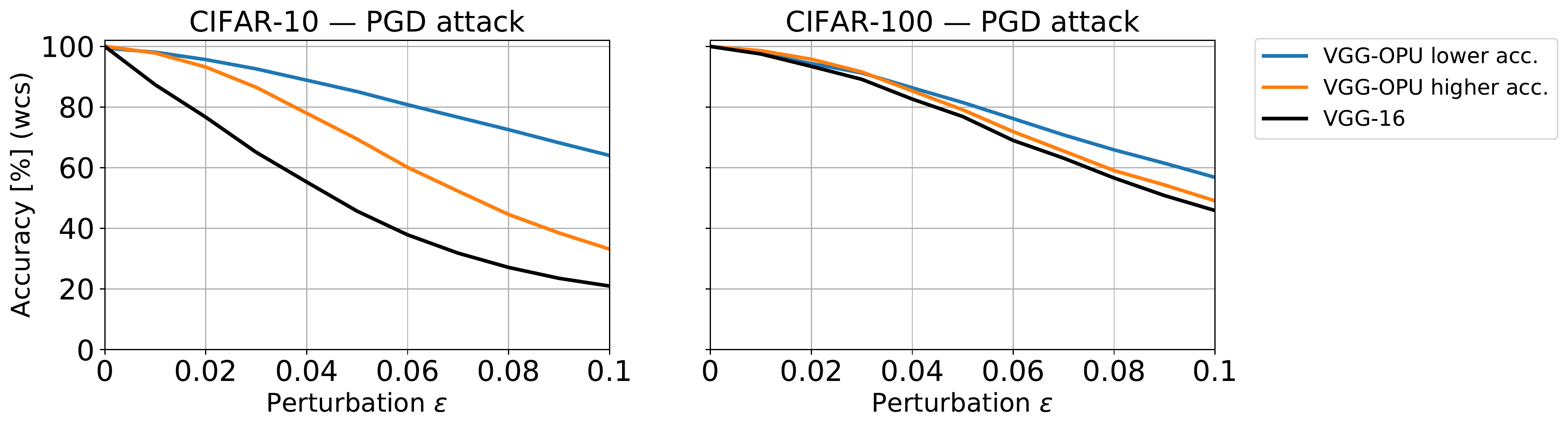}
    \caption{Accuracy on a common well-classified set under transfer attacks of increasing strength. Both VGG-OPU networks are more robust than the the baseline, and we observe a trade-off between robustness and natural accuracy with our defense. Higher/lower accuracies are $\sim85\%$/$80\%$ for CIFAR10 and $\sim72\%$/$68\%$ for CIFAR100. The baseline VGG-16 was trained to reach the higher accuracy.}
    \label{fig:transfer result}
\end{figure}

The target networks are a vanilla VGG-16 trained with backpropagation, and a VGG-OPU optimized as follows: we first train a vanilla VGG-16 with our hybrid training method and then place our OPU defense after the trained convolutional stack, finetuning only the classifier layer.

We also study two different VGG-OPU models, with identical architectures, but different overall natural accuracy to evaluate if there exists a robustness-accuracy trade-off.

\paragraph{Results} The results of this study are shown in Figure \ref{fig:transfer result}. The VGG-OPU network performs better against transfer attacks for all perturbation strength values. On CIFAR-10, we find our defense to provide significant robustness to transfer, between $+15\%$ and $+45\%$ of robustness on the well classified evaluation dataset. This gain is smaller on CIFAR-100, between $+3\%$ and $+10\%$ robustness, where the transfer attack is overall less effective. In either case, we find that there exists an accuracy-robustness trade-off: at the cost of $5\%$ of natural accuracy, robustness can be increased. This makes the defense customizable, allowing to trade some accuracy for robustness. 

\section{Beyond obfuscated parameters: understanding black-box and transfer robustness}\label{section:ablation}
While obfuscated parameters provide robustness by design against white-box attacks, their use alone cannot explain robustness in black-box and transfer scenarios. We perform ablation studies in these two scenarios to understand the increased robustness. We dissect our defense into three different parts: the hybrid training method involving Direct Feedback Alignment \textbf{(DFA)}, the binarization before the optical transform \textbf{(BIN)}, and the non-linear random projection of Equation \ref{eq: opu operation} performed by the OPU \textbf{(RP).} We denote the Optical Processing Unit \textbf{(OPU)} as the combination of \textbf{BIN + RP.} All the models considered are trained to reach similar test accuracy on CIFAR-10 using the hybrid BP-DFA training, to avoid any robustness-accuracy trade-off effect.

\subsection{Black-box attacks}
\begin{figure}[!h]
    \centering
    \includegraphics[width= 0.95\linewidth]{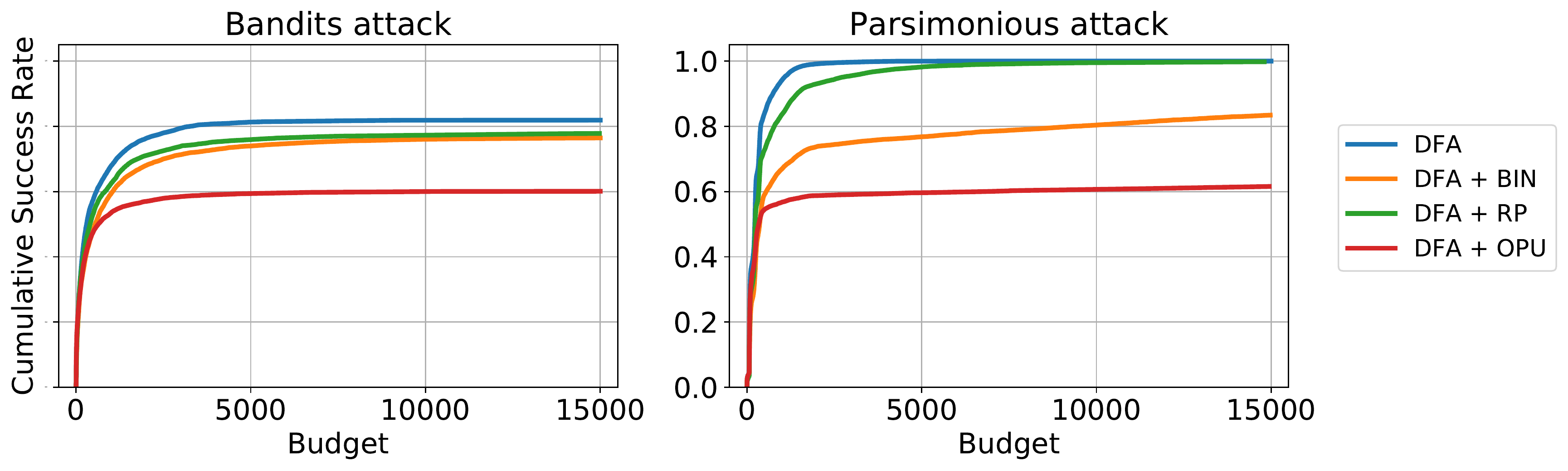}
    \caption{Ablation study on the bandits and the parsimonious attack. The robustness of our defense comes from the binarization and random projection steps. In particular, it's the combination of these two mechanisms that leads to a significant increase in robustness.}
    \label{fig: BB_ablation}
\end{figure}

We first consider an ablation study on a black-box gradient-estimation bandits attacks, in  Figure \ref{fig: BB_ablation}.\\ 
We notice that DFA alone does not provide robustness against gradient-estimation black-box attacks, but adding either a binary layer (BIN) or a random projection (RP) improves robustness by around 5\%. The most striking contribution to the robustness comes from combining the binary layer and the random projection, i.e. the OPU layer, which drops the CSR by 20\%, at no cost in natural accuracy. The failure of DFA to provide robustness against gradient-estimation black-box attack is not surprising: as a training method, it approximates backpropagation with some added noise to the gradients -- this is of little effect to the end model in terms of robustness to gradient-estimation attacks. Conversely, the use of a binarization layer and of random projection change the optimization landscape of the model. In particular, the  binary layer makes this landscape harder to navigate, even for a black-box optimization method.

We then consider an ablation study on parsimonious attacks, as they are based on combinatorial optimization instead of gradient-estimation. The results are shown in Figure \ref{fig: BB_ablation}. We find that the random projection almost doesn't contribute to the robustness. However, binarization has a much stronger effect, decreasing the asymptotic CSR by 20\%. Used in conjunction, the random projection and the binarization decrease the CSR by 40\%. Overall, this confirms that the combination of all the elements of our defense is necessary to obtain black-box robustness. 


\subsection{Transfer attacks}\label{subsec:ablation ta}
\begin{figure}[!h]
    \centering
    \includegraphics[width=\linewidth]{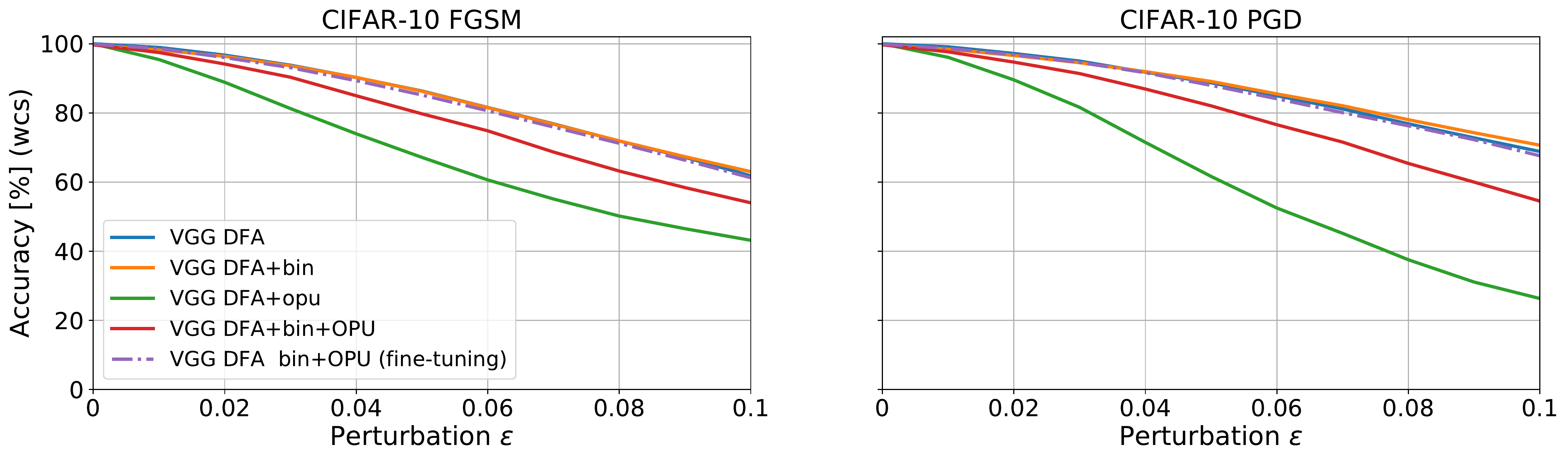}
    \caption{Ablation study of robustness to transfer attacks. All of the architectures are trained to reach similar accuracy with the BP-DFA algorithm. DFA, a VGG-16 trained with BP-DFA is robust on its own: this proves that the hybrid training algorithm is providing robustness against transfer-attacks. Adding a binary layer is not detrimental to the robustness while adding a random projection decreases it. Adding a binary layer before the random projections achieves a compromise. Finally, we show how it is possible to fine-tune a classifier with an OPU on top of the robust features learned by DFA with no loss in robustness and natural accuracy.}
    \label{fig: TA ablation result}
\end{figure}

To complete our ablation study, we seek to understand the source of the robustness against transfer attacks.  Results are plotted in Figure \ref{fig: TA ablation result}. Surprisingly, at variance with our results on classic black-box attacks, the hybrid training method alone is responsible for the robustness against transfer attacks. In fact, the model with hybrid training and OPU is less robust than the model with hybrid training alone. Nevertheless, we find that the robust features learned by hybrid training are transferable: when fine-tuning a classifier including our optical defense on said features, we find we conserve their robustness. We hypothesize that hybrid training with DFA builds features that find a different optimum in the loss landscape than features derived from BP. Accordingly, attacks on features learned by BP transfer poorly to these DFA features. 

\section{Conclusion and outlooks}
We introduced a new defense technique against white-box, black-box and transfer attacks, based on the analog implementation of a neural network layer using an optical co-processor. Our method incurs no additional computational cost at training time, and comes at no natural accuracy cost. We evaluated our method in each setting on the CIFAR-10 and CIFAR-100 datasets, against various attacks. In the white-box setting, our defense is robust by design thanks to \textit{parameter obfuscation}. We attempted to adapt white-box attacks to break this defense, testing two different differentiable approximations, however both resulted in less convincing adversarial examples. Furthermore, in the black-box setting, our defense improves robustness by $22\%$ against the strongest black-box attack that we tested. We also showed increased robustness to transfer of attacks, and showed that there exists in this instance a robustness-accuracy trade-off.

To understand robustness to black-box and transfer of attacks, which cannot be explained by obfuscation, we performed an ablation study of our defense. In black-box scenarios, we found that the combined use of a random projection and binarization significantly strengthen our defense, highlighting how the OPU is at the center-stage of our defense.  In transfer of attacks, surprisingly, we found that the hybrid training method we devised to train neural networks incorporating our defense builds robust features, enabling our defense layer to be robust against transfer of attacks. These findings motivate further studies on the loss landscape explored by alternative training methods. In essence, we find that every aspect of our strategy, from the analog optical co-processor and its binarized input, to the hybrid training method we use, contribute in different settings to building an all-around robust defense. In particular, while our defense can be simulated, its physical implementation by an optical co-processor guarantees the parameters remain obfuscated even if the host system is compromised -- a unique feature. 

Future work could investigate the combination of our defense with other techniques based on ensembling or adversarial training. Finally, we note that attackers succeeding in efficiently breaking our obfuscation of parameters could have a significant impact in imaging and phase retrieval.
\section*{Acknowledgments}
The authors thanks the LightOn team for their useful feedback. RO acknowledges support from the Région Ile-de-France. FK acknowledges support by the French Agence Nationale de la Recherche under grant ANR17-CE23-0023-01 PAIL and ANR-19-P3IA-0001 PRAIRIE. Additional funding is acknowledged from “Chaire de recherche sur les modèles et sciences des données”, Fondation CFM pour la Recherche-ENS.

{\small
\bibliography{bibliography}
}

\appendix
\newpage
\setcounter{figure}{0} 
\renewcommand{\thefigure}{S\arabic{figure}}
\section{FGSM supplementary results}
\begin{figure*}[h]
    \centering
    \includegraphics[width=0.95\linewidth]{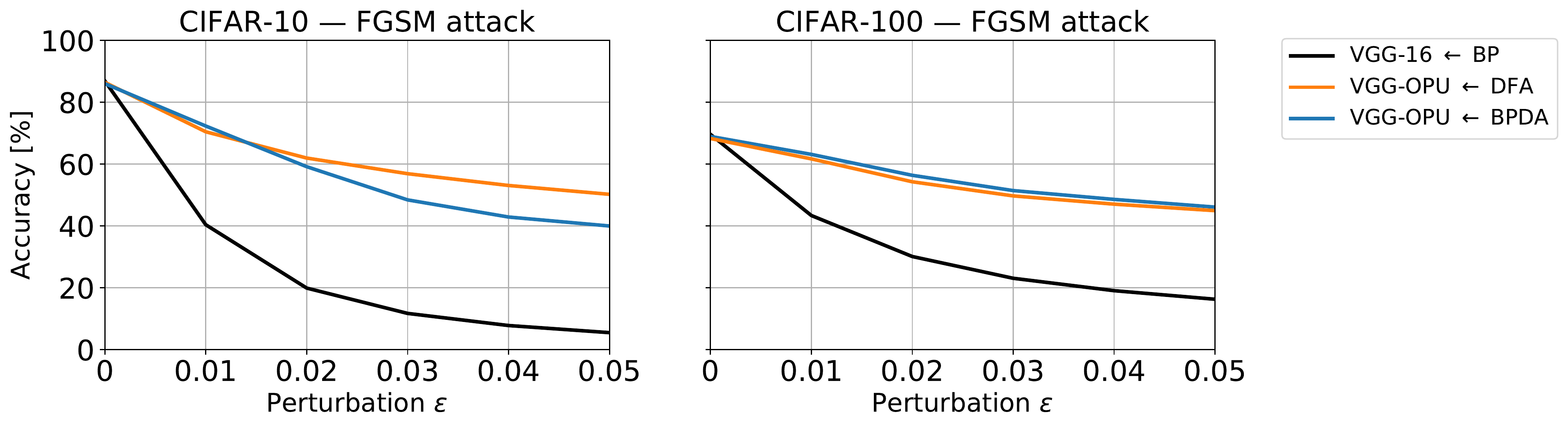}
    \caption{Notation: $<$model$>$ $\leftarrow$ $<$attack gradients$>$. For example VGG $\leftarrow$ BP means that a VGG-16 is attacked with gradients computed with backpropagation. The VGG-OPU model is again more robust than the VGG-16 baseline. BPDA fails again to produce better attacks than our hybrid training method.}
    \label{figsup: WB attack result}
\end{figure*}

\begin{figure*}[h]
    \centering
    \includegraphics[width=0.95\linewidth]{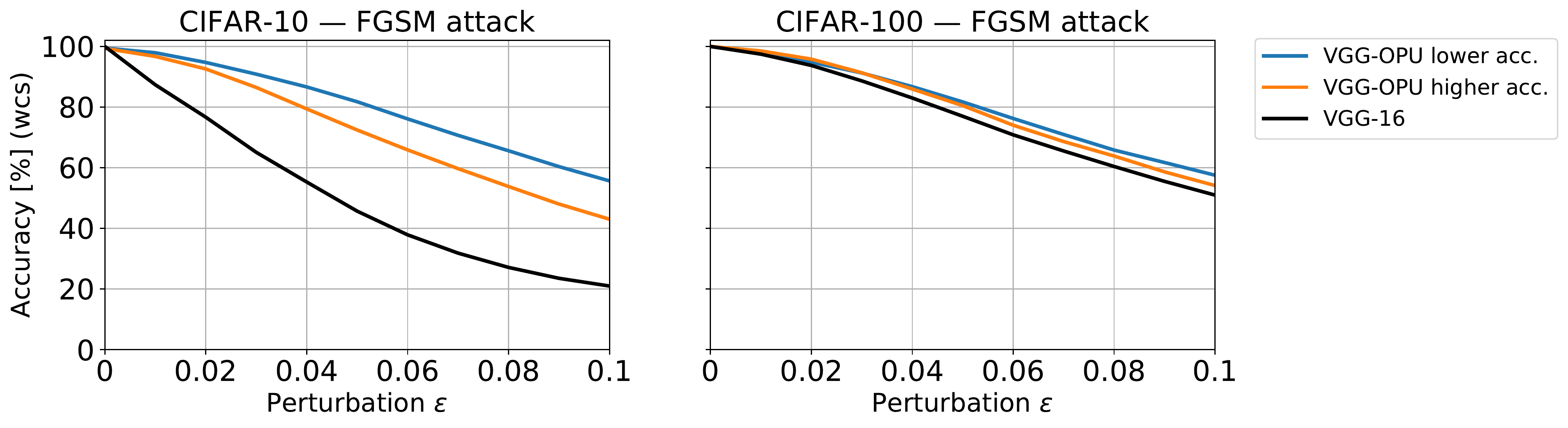}
    \caption{Accuracy on a common well-classified set under transfer attacks of increasing strength. Even for FGSM attacks both VGG-OPU networks are more robust than the the baseline. Once again we observe a trade-off between robustness and natural accuracy with our defense. Higher/lower accuracies are $\sim85\%$/$80\%$ for CIFAR-10 and $\sim72\%$/$68\%$ for CIFAR-100. The baseline VGG-16 was trained to reach the higher accuracy.}
    \label{figsup:transfer result}
\end{figure*}

Figures \ref{figsup: WB attack result} and \ref{figsup:transfer result} show 
how the results for white-box and transfer attacks also hold for FGSM attacks. 
\section{Description of our VGG models}

The baseline architecture we use is a Vanilla VGG-16. The convolutional stack and the classifier are based on \url{https://github.com/chengyangfu/pytorch-vgg-cifar10} (configuration D with batchnorm). For CIFAR-100 we simply modify the last linear layer to adapt to the higher number of classes. 

The Vanilla VGG-16 has three fully connected (fc) layers after the convolutional stack of sizes: fc1: $512-512$, fc2: $512-512$, fc3: $512-\textrm{classes}$ ($\textrm{classes}=10$ for CIFAR-10 and $100$ for CIFAR-100). The VGG-OPU shares the same convolutional stack of the vanilla VGG-16; it has however a simulated or real OPU (binary layer followed by a nonlinear random projection) instead of fc2. We have used OPU input size $512$ and output size $8000$ for white-box attacks, input size $1024$ and output size $8000$ for transfer attacks and the ablations study, finally input size $2000$ and output size $10000$ for black-box attacks.
Further details on the hyperparameters used to train the models (e.g. learning rate, weight decay) can be found in the code. 

\section{Adversarial attacks hyperparameters}
\label{sectionsup: hyperparameters}
\textbf{White-box attacks} were performed on images normalized in $\left [-1, 1 \right ]$. To create adversarial attack with PGD we used $50$ iterations with $\alpha = 0.01$ step size. The PGD and FGSM attacks for transfer experiments have been created using the same parameters. 

\textbf{Black-box} attacks in the following were performed on images normalized in $\left [0, 1 \right ]$.

\textbf{NES attacks} were performed using the cross-entropy loss on the $L_\infty$ ball of radius $8/256$. The maximum number of queries is $15000$, the number of samples to estimate the gradients is $50$ and the size of a batch 1024. We vary the standard deviation $\sigma$ at values $\left[0.05, 0.1, 0.5, 1\right]$. We keep the attack with the $\sigma$ yielding the best Cumulative Success Rate for each model on each dataset.

\textbf{Bandits attacks} were also performed using the cross-entropy loss on the $L_\infty$ ball of radius $8/256$. The maximum number of queries is $15000$ and the number of gradient iterations is kept to $1$. We vary the standard deviation $\sigma$ at values $\left[0.1, 0.5, 1\right]$. The online learning rate is set at $0.1$, the exploration at $0.1$ and the prior size at $16$.

\textbf{Parsimonious attacks} were performed with $\epsilon=8/256$, the number of iterations in local search is $1$, the initial block size is $4$, the size of a batch is $64$ and no hierarchical evaluation was performed.


\section{Retrieval of obfuscated parameters}
\label{sectionsup: retrieval}
\paragraph{Setting} With direct access to the host system and the optical co-processor, it is possible to use phase retrieval techniques to retrieve the transmission matrix of the co-processor and hence the obfuscated parameters. 

\paragraph{Setup} The timings mentioned in the paper are obtained on a server with a dual CPU setup, with Intel(R) Xeon(R) Gold 6130 CPU @ 2.10 GHz. Implementing the same algorithm on GPU has only a marginal effect on performance.

\paragraph{Limitations} The complexity scales linearly with the output size, $O(n \log n)$ with the input size $n$, and quadratically with the relative error, that is reducing the relative error by half costs a $4 \times$ increase in execution time. However, the main limit of the retrieval algorithm is in terms of memory use and co-processor abilities: to recover a transmission matrix with $10^3$ rows/columns, we need to project and process a matrix with $10^5$ rows/columns (the calibration signal). First, this procedure rapidly requires large amounts of RAM. Moreover, given the limit of input and output size of the optical system at $10^6$, the impossibility to increase the calibration signals further limits the achievable precision of the retrieval.

Finally, it is possible to change the transmission matrix of the optical system, simply by changing some characteristics of the input light (e.g. the wavelength).

\end{document}